\pdfoutput=1
\documentclass{article}
\usepackage[square,sort,comma,numbers]{natbib}
\usepackage{arxiv}

\usepackage[utf8]{inputenc} %
\usepackage[T1]{fontenc}    %
\usepackage{hyperref}       %
\usepackage{url}            %
\usepackage{booktabs}       %
\usepackage{amsfonts}       %
\usepackage{nicefrac}       %
\usepackage{microtype}      %
\usepackage{lipsum}
\usepackage{amsmath}
\usepackage{graphicx}
\usepackage{todonotes}
\usepackage{wrapfig}
\usepackage[toc,page]{appendix}

\title{Multi-node Bert-pretraining: Cost-efficient Approach}

\author{Jiahuang Lin\\University of Toronto\\Vector Institute\\\texttt{jacoblin@cs.toronto.edu}
\and Xin Li\\University of Toronto\\ Vector Institute\\\texttt{xin.li@vectorinstitute.ai}
\and Gennady Pekhimenko\\University of Toronto\\Vector Institute\\\texttt{pekhimenko@cs.toronto.edu}}

\begin{document}
\maketitle
\begin{abstract}
Recently, large scale Transformer-based language models such as BERT, GPT-2, and XLNet have brought about exciting leaps in state-of-the-art results for many Natural Language Processing (NLP) tasks. One of the common trends in these recent models is a significant increase in model complexity, which introduces both more weights and computation. Moreover, with the advent of large-scale unsupervised datasets, training time is further extended due to the increased amount of data samples within a single training epoch.  As a result, to train these models within reasonable time, machine learning (ML) programmers often require advanced hardware setups such as the premium GPU-enabled NVIDIA DGX workstations or specialized accelerators such as Google's TPU Pods. Our work addresses this limitation and demonstrates that the BERT pre-trained model can be trained within 2 weeks on an academic-size cluster of widely available GPUs through careful algorithmic and software optimizations. In this paper, we present these optimizations on how to improve single device training throughput, distribute the training workload over multiple nodes and GPUs, and overcome the communication bottleneck introduced by the large data exchanges over the network. We show that we are able to perform pre-training on BERT within a reasonable time budget (12 days) in an academic setting, but with a much less expensive and less aggressive hardware resource requirement than in previously demonstrated industrial settings based on NVIDIA DGX machines or Google's TPU Pods.
\end{abstract}

\section{Introduction}
The BERT language model~\cite{bert} has significantly improved the state-of-the-art performance of many downstream NLP tasks such as language understanding and question answering. However, training BERT is computationally intensive due to its high model complexity and large amount of training data that needs to be processed to achieve the state-of-the-art model accuracy. There are 110M parameters in BERT base and 340M parameters in BERT large, and, due to the unsupervised nature of the algorithm, the model is trained on an abundance of unlabeled data over many epochs, e.g., the BookCorpus dataset~\cite{bookcorpus} (800M words) and the English Wikipedia (2500M words) dataset were trained for 40 epochs in the original BERT model~\cite{bert} . Nonetheless, with BERT being a Transformer-based language model~\cite{transformer}, the stacked attention and fully-connected layers allow significantly more parallelism as compared to sequential models (e.g., LSTM RNN models~\cite{lstm, rnn1, rnn2, rnn3, rnn4}), as LSTM RNNs have a higher memory requirement that prevents the GPU to utilize its compute cores efficiently~\cite{echo}. This architectural advantage of BERT naturally provides opportunities to train BERT on multiple GPU/TPU devices, which reduces the training time linearly with the amount of hardware resource available. 

Nonetheless, machines that are capable of servicing such level of parallelism are usually very expensive and overly dedicated to computationally intensive workloads ~\cite{expensive_hardware}. For example, NVIDIA released its BERT pre-training results that took place on 32 interconnected DGX-1, DGX-2, or DGX-A100 workstations~\cite{bertimplementation}. The DGX machines are equipped with high-end GPUs, CPUs, and interconnects~\cite{dgx}, and have a unit price ranging from \$150K to \$400K ~\cite{dgx_price}. Therefore, such a hardware setup requires an initial capital cost of around \$4.8M to \$12.8M. The capital cost to enable such experiment is not affordable for many researchers and research institutions despite the significant improvement in training time for the large models. In addition, the setup is over-powered to the routine computation needs of many typical ML users. For example, Vector Institute only averages 1.28 GPUs per job allocation over a month period for one of its major clusters equipped with 64 nodes, 8 GPUs each. This indicates a relatively low interest in academic users to utilize the high-speed interconnects and communication collectives provided by the system, which is unsuitable for the configuration of DGX machines and other high-performance workstations alike. Anecdotally, we also observe relatively low usage of advanced hardware features such as TensorCores~\cite{tensorcore} and mixed precision~\cite{mixed_precision}. We attribute this to two factors: (i) limited awareness among ML researchers/practitioners into advanced hardware features for ML and (ii) the necessity to deal with problems such overflow/underflow in FP16 computation~\cite{mixed_precision, mixedprecision_graph_rewrite} that require special software tricks~\cite{mixedprecision_graph_rewrite} to be handled properly.

Our work aims to exploit the same parallelism in BERT model, but distribute the workload over a much cheaper 32-node cluster available to us at Vector Institute connected with a commodity 10Gb/s network, where each node is equipped with 8 low-budget NVIDIA T4 GPUs~\cite{nvt4}. We apply multiple layers of optimization to improve both the single GPU performance and distributed performance over the network and over internal PCIe interconnect (used within a node). With a much lower hardware budget of approximately \$600K and a much more generic compute environment, we are able to achieve a 70\% weak scaling efficiency and complete BERT training in 12 days, which we consider to be a reasonable training time under an academic setting.

\section{Related work}
\label{sec:headings}

The pre-trained language models such as BERT~\cite{bert}, GPT-2~\cite{gpt2}, XLNet ~\cite{xlnet}, RoBERTa~\cite{roberta} and GPT-3~\cite{gpt3} have been proven successful on many downstream NLP tasks including text classification, question answering, and natural language inference. In this paper, we will focus on the pre-training of BERT-large. We pick BERT-large because of its wide adoption and state-of-the-art performance. From a systems perspective, BERT-large is also a suitable candidate since the encoder based attention layers~\cite{transformer}` in BERT have similar characteristics to many of the aforementioned models, and is considered as a next generation ML benchmark for the state-of-the-art ML benchmark suites such as MLPerf~\cite{mlperf}.  

\subsection{BERT Pre-Training}
In the original BERT paper~\cite{bert}, the model reuses the encoder implementation of the Transformer model~\cite{transformer}. Pre-training is done with two tasks: (i) the masked language model, where the model predicts words that are randomly masked in the input sentence, and (ii) the next sentence prediction, where the model need to classify whether two input sentences are logically adjacent. A combination of Wikipedia Corpus and BooksCorpus dataset is used to train BERT-large. The pre-training processes is carried out into two separate phases where phase 1 covers the first 90\% of epochs using a sequence length of 128 to improve training speed, and phase 2 covers the rest 10\% of epochs with a sequence length of 512 to learn the positional embeddings. The two training phases add up to a total of 40 epochs and take over 80 hours to complete on a 64 TPUv3 chips~\cite{bert}. 

To reduce training time, a natural choice is to increase the training mini-batch size. A larger mini-batch size (number of samples used for each back-propagation update) decreases the total number of iterations per training epoch as the number of total samples in the dataset is fixed. Although a larger mini-batch size also increases the computation load to each iteration, this additional workload can be fully parallelized given enough computation resources available. Nonetheless, experience has shown that learning rate will need to be carefully fine-tuned for large mini-batch learning to be successful~\cite{imagenet}. As a result, fairly large mini-batch has prevented pre-training from continuing to scale out with more hardware resources. To address the aforementioned issue, LAMB optimizer~\cite{lamb} was introduced, where the gradients and learning rate will be further normalized and dynamically adjusted in a layer-wise manner. The LAMB optimizer paper has shown that by following the two phases pre-training convention, BERT-large pre-training time can be shorten to 76 minutes by scaling to 1024 TPUv3 chips~\cite{lamb} without a loss of model accuracy. 

\subsection{Distributed Training}

As deep learning models become more powerful and complex, the training of those models also demands more computation resources. Large models like ResNet~\cite{resnet} and DeepSpeech2~\cite{deepspeech2} can take weeks to train on a single GPU device~\cite{openaicompute}. The need for shortening the training time of large deep learning models has brought up distributed training algorithms. Among the distributed training algorithms, data parallelism~\cite{dataparallel} and model parallelism~\cite{modelparallel} are the two most popular types. Data parallelism~\cite{dataparallel} is a natural way to scale out the training process by slicing and distributing the training data into multiple devices. Each worker will retain a full replica of the model on different data. Workers will synchronize over the updates of the model by exchanging gradients. In contrast, model parallelism~\cite{modelparallel} divides the model into different pieces and distributes those pieces into each devices to form a training pipeline. Workers will train on the same date but for different parts of the model, which allows devices to fit bigger model. Workers will synchronize over the activation maps.

However, as the training graphs of deep learning models are typically directed-acyclic, model parallelism~\cite{modelparallel}, which partitions the execution graph, essentially introduces strong sequential dependencies: at most one device will be fully utilized in computing at any given time of training. To gain more device utilization, pipeline devices to overlap computation was proposed~\cite{pipelineparallel}. Such overlapping requires activation maps that are supposed to be synchronized to be stacked and stored. Assuming mini-batch data enters the pipeline one at a time, and the pipeline has length $n$, this will impose an extra linear scale of memory storage for devices in the pipeline, which is not scalable as the number of devices increases. Furthermore, the hard limit on device memory that model parallelism~\cite{modelparallel} brings leaves very little room for researchers to optimize the training throughput. Compared with model parallelism~\cite{modelparallel}, data parallelism~\cite{dataparallel} also introduces hard limit: each device must be able to fit in one complete replica of the model. One might argue that this is a bottleneck as models are getting bigger, it's the feature maps that consumes most memory rather than the model itself. Although model size usually is not a bottleneck, data parallelism suffers from trade-offs between synchronization cost and model parameter staleness in parameter updates~\cite{dataparallel}.

Network topologies have also been explored in the implementation of data parallelism in order to reduce the synchronization cost. For example, a ring based system topology~\cite{nccl} has been proposed to maximize the inter-device communication bandwidth. By having all devices jointly form a ring topology, each device only passes the computed weight gradient to its neighbor in the ring. Such approach guarantees that communication channel between any two devices will be filled up with maximum one model's gradient, avoiding traffic congestions. Further benchmarks have shown this approach guarantees linear scalability of bandwidth with respect to the number of devices~\cite{nccltest}.

\subsection{Mixed Precision Training}

Reducing the arithmetic complexity can be beneficial to the run-time performance of training procedures, especially for large models, as the weights of deep learning model consume a great amount of memory. In addition to the reduction of memory footprint, lower precision number representations also reduce the numerical computation complexity, and thus increases the calculation throughput. This effectively shortens the program execution time.  For example, deep learning models usually use full precision floating point numbers (FP32) to store the weights and carry out computations. The work on mixed precision training~\cite{mixed_precision} showed the possibility of using 16-bit floating point numbers (FP16) while preserving similar convergence behavior and model performance for DNNs. During training, FP16 are used to perform multiplications between the weights and activations, and FP32 are used for the accumulation of the products during the reduction. Since multiplications require more hardware resource, this optimization can significantly improve the compute efficiency during training. Moreover, with the introduction of TensorCore~\cite{volta_gpu}, direct hardware support is provided to this mixed precision multiplication and accumulate pattern. As a result, mixed precision training can improve the training throughput and shorten the training time by 2-6 times on various representative DNN models~\cite{mixed_precision}. This migration also effectively reduces the memory footprint during training, which allows for larger batch size to fit in GPU memory.

Loss scaling is used to compensate for the loss of dynamic range from FP32 to FP16. During training, the gradients usually have a very small magnitude (negative exponent). Since the exponent bits in FP16 have a representation range of [-14,15], most of the positive exponent range is left unused while many small-magnitude gradients are rounded to zero. To mitigate the zeroing of the gradients during the backward pass, the gradients are scaled up by a constant factor to take advantage of the unused range of positive exponents. They are then scaled down before the weight updates to preserve the same update magnitude to the original FP32 model. 

\section{Methodology}
We explain our training setup and distribution strategies used to train the BERT-large model below. 
\subsection{Datasets}

\subsubsection{Pre-training Datasets}

Similarly to the original BERT paper~\cite{bert}, we also used Wikipedia Corpus~\cite{wikiextractor} and BookCorpus~\cite{bookcorpus} dataset. The Wikipedia Corpus has 2.5B words and BooksCorpus has 800M words. After extracting plain English text from those two public datasets, we then process the sentences exactly like in BERT~\cite{bert}. Namely, 
\begin{itemize}
    \item tokenize the raw text through WordPiece tokenization~\cite{wordpiece}
    \item mask out 15\% of the words in the input sentences for the model to learn the relationship within the sentence
    \item split and shuffle adjacent sentences with 50\% probability for the model to do next sentence prediction
\end{itemize}

\subsubsection{Fine-tuning Datasets}

For fine-tuning, we picked the the question-answering task trained on the Stanford Question Answering Dataset (SQuAD 1.0)~\cite{squad}. This dataset contains 100k question-answer pairs from Wikipedia, in which a question and a passage is provided as the training data and the corresponding answer is provided as the training label.

\subsection{System Setup}
\begin{figure}
    \centering
    \includegraphics[width=0.6\textwidth]{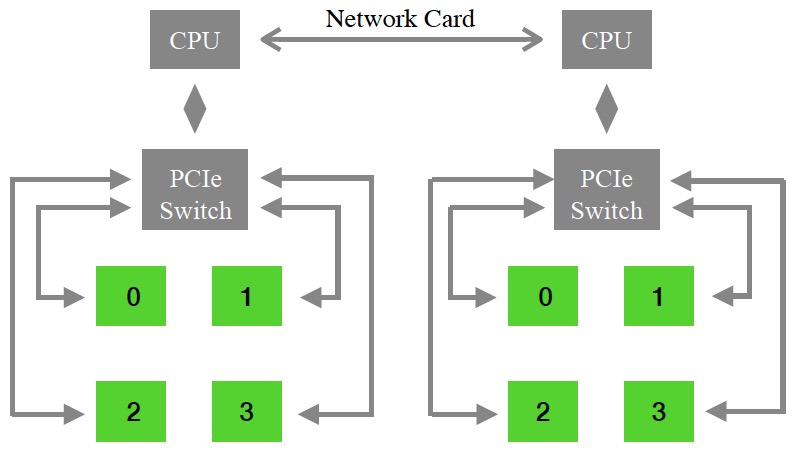}
    \caption{PCIe based Data Parallelism}
    \label{fig:fig1}
\end{figure}

We utilized data parallelism~\cite{dataparallel} for multi-node training. To characterize the hardware topology, we use the name "<X>M<Y>G" to encode X number of "Machines" and "Y" number of "GPUs" on each machine. Table~\ref{tab:sys} shows our 32M8G hardware setup to conduct training. This setup has a unit cost of \$19.5K  per node, which is much less expensive than the unit price of a single DGX system, which ranges from \$150K to \$400K~\cite{dgx_price}. The system also offers more flexibility because it is more cost effective for light-weight computing needs, which do not utilize the interconnects as heavily as is required in distributed training for some large models. For simplicity, let us consider a simple scenario of a 2-node, 4 GPUs each, setup as shown in Figure~\ref{fig:fig1}. Each node will have four NVIDIA T4 cards in our case and are connected through PCIe. Nodes will be connected through network card.

\begin{table}
 \caption{Multi-node Hardware Setup for BERT-large Training}
  \centering
  \begin{tabular}{llll}
    \toprule
    System Requirements         &   Value \\
    \midrule
    Node Count                  & 32 \\
    GPU Per Node                & 8  (NVIDIA T4) \\
    CPU Per Node                & 32 (Intel(R) Xeon(R) Silver 4110 CPU @ 2.10GHz) \\
    Total CPU count             & 1024 \\
    Total GPU count             & 256 \\
    GPU-Interconnect            & PCIe 4.0 link with 64Gb/s bandwidth \\
    Network Connection Between Nodes            &  10 Gb/s. \\
    Estimated Cost of Acquisition Per Node & \$19.5K \\
    Estimated Total Cost of Acquisition & \$624K \\
    \bottomrule
  \end{tabular}

  \label{tab:sys}
\end{table}

For the software stack, we used NVIDIA's PyTorch implementation of the BERT model~\cite{bertimplementation}. Table~\ref{tab:software} shows the software stack of our training program. We also utilized NCCL v2.4.8-1~\cite{nccl} as our distributed training backend framework that implements data parallelism~\cite{dataparallel}. In addition, NCCL~\cite{nccl} also auto detects the network topology upon setting up connections among nodes in order to form a ring topology if possible~\cite{nccl}.

\begin{table}
 \caption{Multi-node Software Setup for BERT-large Training}
  \centering
  \begin{tabular}{llll}
    \toprule
    Major Software Requirements         &   Version \\
    \midrule
    Ubuntu              & 18.04 \\
    CUDA                & 10.0 \\
    cuDNN               & 7.5.1 \\
    Python              & 3.6 \\
    PyTorch             & 1.1.0 \\
    APEX (A Pytorch Extension)                & 0.1 \\
    NCCL                & v2.4.8-1 \\
    MKL                 & 2019.4 \\
    \bottomrule
  \end{tabular}

  \label{tab:software}
\end{table}

When doing training in the context of data parallelism~\cite{dataparallel}, mini-batches of data are split into individual devices. In the forward pass, the calculated gradients will be collected. Gradients will be exchanged through PCIe within a node and through network card across different nodes, after which the parametres are updated based on the aggregated results. In particular, the PCIe gradient pass is independent with the network card gradient pass. During the backward pass, the whole model's gradients in any device will need to be passed to all other devices as each device contains a full replica of the model in data parallelism. As such, we can tell from the previous bandwidth comparison that the network gradient transmission will certainly lag behind the PCIe gradient transmission. And we will see later that this is indeed the case: the time needed for completing a backward pass is dominated by the gradient exchange time spent over the network.

\subsection{Pre-training}
We trained BERT-large on the combined Wikipedia Corpus and BooksCorpus dataset. We also followed the two phase training schema when performing pre-training ~\cite{bert}), where we used a sequence length of 128 to train the first 36 epochs and a sequence length of 512 to train the last 4 epochs.

\section{Optimization}
\label{sec:optiomation}
To improve the single device throughput and fine-tune the training workload with respect to our distributed system topology, we performed the following optimizations on BERT-large pre-training.

\subsection{Data Sharding}
Considering the significant size of raw data: 2.5B words from Wikipedia Corpus~\cite{wikiextractor} and 800M words from BooksCorpus~\cite{bookcorpus}, file I/O became a bottleneck during data loading. Due to the essence of data parallelism~\cite{dataparallel}, the data file is loaded into memory first for each node. And then the data will be truncated and each device will get a portion of it. However, with this large dataset, data loading and distributing brings long latency upon the start of each epoch training. Our benchmark result shows that it will take 8 to 10 minutes to load and distribute data in a single machine that has 8 GPUs when the training program freshly starts. This latency will be shortened to 3 to 5 minutes later upon epoch re-shuffling and re-distributing.

To address this I/O bottleneck, we perform data sharding before training based on the number of devices we have. During training, each device will only read from the data shard that it belongs to in order to avoid the I/O congestion caused by data transferring. Each processed shard is stored in hdf5 file format~\cite{hdf5} for flexible data storage and efficient I/O during distributed training. Those shards are evenly distributed segments of the original dataset. This organization of data facilitates distributed training because data can be dispatched and shuffled efficiently to different nodes in the network. After data sharding, our results shows that data loading time upon the start of the training program has decreased to less than 2 minutes, and the time spent on the re-loading in the beginning of each epoch is less than one minute.

\subsection{Automated Mixed Precision}
During pre-training, the BERT model stores the weights as 32-bit full precision floating point numbers (FP32), which means the training is carried out with 32-bit multiplication and accumulation compute units. We applied automated mixed precision~\cite{mixed_precision} to the pre-training of BERT-large. In other words, we convert the model to use half precision data type whenever possible while keeping a full precision model weights replica in the master node. Loss scaling~\cite{mixed_precision} is used to preserve the small gradient values. Depending on the device, empirical benchmark result \ref{tab:table1} shows 1.7$\times$ to 2.5$\times$ speedups after applying automated mixed precision for the pre-training of BERT-large.

Typically in a computation graph, not all FP16 operators are numerically safe. This means operators that are considered numerically dangerous will have its calculation in full precision. For example, a \emph{plus} operator is marked as safe while a \emph{power} or a \emph{log} operator is considered numerically dangerous in half precision. Automated mixed precision handles the categorization of the numerical safety level through the rewriting of computation graph~\cite{mixedprecision_graph_rewrite}.

\subsection{Kernel Fusion}
A CUDA kernel is a compiled routine that runs on NVIDIA GPUs. These kernels are highly optimized to perform matrix algebra operations in the model. Although PyTorch uses Python as its front-end interface for users to build up their deep learning applications, to speed up training, the front-end Python interpreter will invoke compiled CUDA kernels to perform training on GPU devices. Generally, kernels are provided for each operator in the front-end language. For example, a sequence of operations, such as a matrix-matrix multiplication followed by an element-wise $tanh$ activation would produce two CUDA kernels corresponding to the two operators. Another example would be Gauussian Error Linear Unit~\cite{gelu}, which is heavily used as an activation function in BERT~\cite{bert}. The GELU function was approximated by the following:
\begin{equation*}
    GELU(x) = 0.5x(1+tanh[\sqrt{2/\pi}(x + 0.044715x^3)])
\end{equation*}
By replacing the constants with $a$, $b$ and $c$ we get:
\begin{equation*}
    GELU(x) = ax(1+tanh(b(x+cx^3)))
\end{equation*}
Without kernel fusion, the above equation will translate into 7 kernels as the following:
\begin{enumerate}
    \item $f = x^3$
    \item $f = c * f$
    \item $f = x + f$
    \item $f = b * f$
    \item $f = tanh(f) + 1$
    \item $f = x * f$
    \item $f = a * f$
\end{enumerate}
This is inefficient compared to a single fused CUDA kernel combining all operators because the fused kernel incurs less kernel launch overhead and the access of the same piece of data exploits better memory locality. Therefore, we applied kernel fusion for both layer normalization~\cite{layer_norm}, activation functions~\cite{gelu} and the optimizer~\cite{lamb} using the Apex utility functions provided by~\cite{apex}. Based on our benchmarking result, the throughput improved by 8\% to 11\% on average depending on the device. 

\subsection{Multiple Node Training}
\begin{figure}[h]
    \centering
    \includegraphics[width=0.52\textwidth]{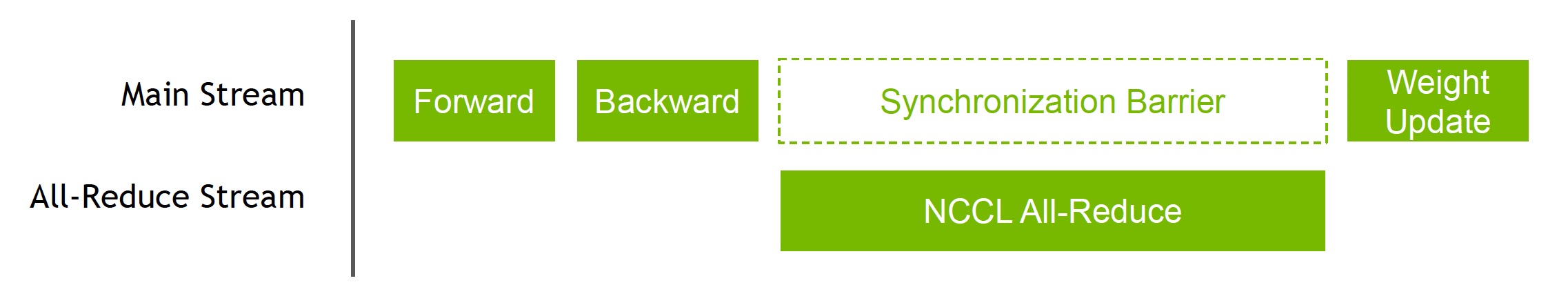}
    \includegraphics[width=0.38\textwidth]{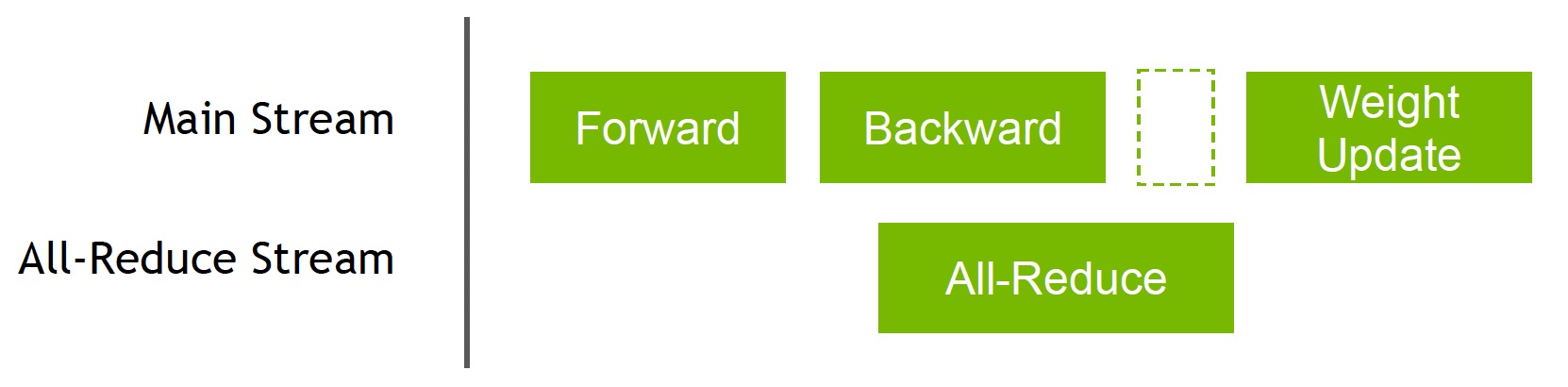}
    \caption{Timeline Comparison Between Non-overlapping/Overlapping Communication with Computation}
    \label{fig:all-reduce-overlapping-comparison}
\end{figure}

While the results for single GPU optimization might seem promising, training BERT-large with a single GPU is still practically infeasible. Table~\ref{tab:table3} justifies the need for multi-node training as it will take years to train BERT-large in a single GPU setting. We thus exploit data parallelism to scale to multiple GPUs in a multi-node context. During training, each GPU has a complete copy of the model, and they are provided with different batch of input training data. Each GPU worker first computes the gradient with respect to its input individually, the gradient are then exchanged and accumulated across different workers through the NCCL ~\cite{nccl}). Since both data loading and weight sharing consumes communication bandwidth of the interconnects, careful scheduling and allocation of the communication bandwidth is required to minimize congestion and maximize GPU utilization. As a result, we perform data loading via the PCIe channels and weight sharing via the network, which minimizes the competition of resources. To improve GPU utilization in the presence of a non-trivial communication workload, we overlap the gradient computation with communication as illustrated in Figure~\ref{fig:all-reduce-overlapping-comparison}. The gradients are exchanged as soon as they become available after passing some certain size threshold during the backward pass, so back-propagation and weight exchange can happen in a parallel in a pipelined fashion. 
\begin{table}[]
    \caption{Single GPU Pre-training Time Estimation}
      \centering
      \begin{tabular}{lllll}
        \toprule

        Device          & Optimized Throughput     & Tokens/Epoch   & Estimated Time Per-Epoch   & 40 Epoch Time\\
        \midrule
        P100~\cite{nvp100}                    & 3228.8/s              & 16752.7 Million       & 1441.6 hours (60 days) & 2400 days\\
        T4(TensorCore)~\cite{nvt4}         & 5429.1/s              & 16752.7 Million       & 857.1 hours (36 days) & 1440 days\\
        2080Ti~\cite{nvrtx2080}                  & 10765.8/s             & 16752.7 Million       & 432.3 hours (18 days) & 720 days\\
        \bottomrule
      \end{tabular}
      \label{tab:table3}
\end{table}

\begin{figure}[h]
    \centering
    \includegraphics[width=0.45\textwidth]{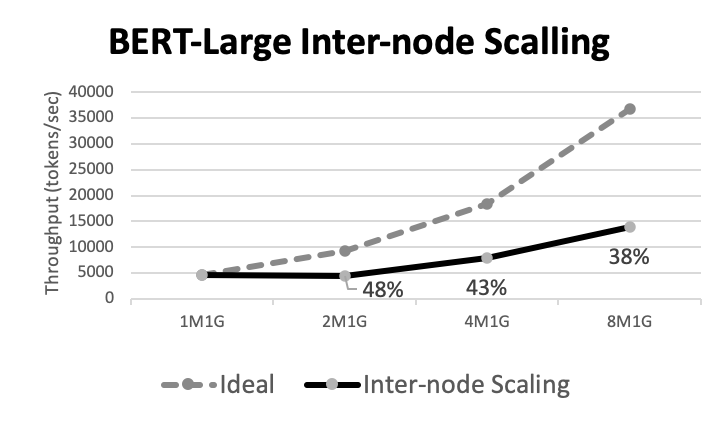}
    \hspace{0.05\textwidth}
    \includegraphics[width=0.45\textwidth]{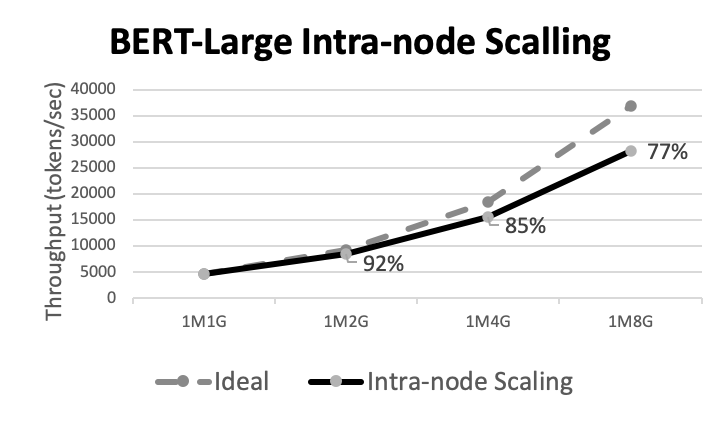}
    \caption{Weak Scaling Comparison Between Intra-node Scaling and Inter-node}
    \label{fig:intranode-vs-internode-scalling}
\end{figure}

There are two types of communication happening during the above process, namely intra-node communication through PCIe and inter-node communication through the network. As our environment has only 10 Gbps network bandwidth, inter-node communication quickly became the bottleneck for multi-node training. Our benchmarking shows that after overlapping communication with computation, in a simple scenario of training on 2 nodes with each node having one GPU respectively, time spent on synchronization barrier is comparable to the forward pass, the backward pass and the weight update pass combined. Figure~\ref{fig:intranode-vs-internode-scalling} further illustrates this observation. In this figure, the X-axis denotes the hardware configuration as we scale up the computation resource for the training.We use the name "<X>M<Y>G" to encode X number of "Machines" and "Y" number of "GPUs" on each machine. For Inter-node scaling, i.e. increasing "X", one can see that there is nearly zero throughput gain going from 1M1G to 2M1G as almost half of the time was spent on the communication rather than computation. In addition, we can see from Figure~\ref{fig:intranode-vs-internode-scalling} that the weak scaling efficacy is upper bounded by 38\% in practice, which is significantly less efficient than the Inter-node scaling alternative.
 
 \begin{wrapfigure}{r}{0.45\textwidth}
    \centering
    \includegraphics[ width=0.3\textwidth]{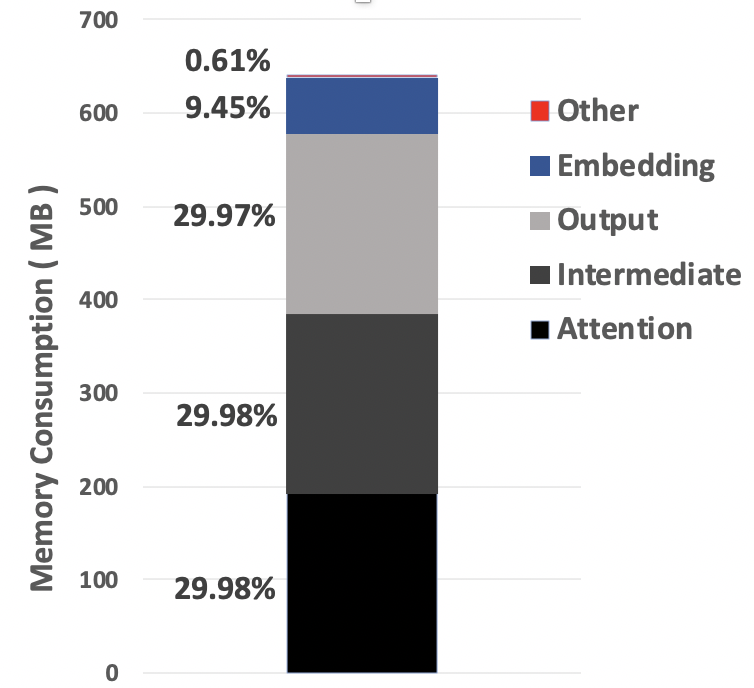}
    \caption{Gradient Memory Profile}
    \label{fig:gradient_mem}
\end{wrapfigure}

With network bandwidth being the hard limit of hardware, reducing the amount of data each node needs to transmit became one solution for maintaining scalability. Prior work~\cite{gradientsparsification} proposed Gradient Sparsification to reduce the size of gradients. However, sparsification is effective primarily in sparse gradient data. This characteristic is unfavorable for our model. As we show in Figure~\ref{fig:gradient_mem}, the majority of the gradients are in the attention, intermediate, and output layers, which are mostly fully-connected layers that produce dense gradients. In addition, picking the right sparsification threshold not only requires an extra amount of calculation overhead but also a lot of tuning work: if the threshold is too low then we are only able to reduce the gradient size by little amount; if the threshold is too high then we are in risk of affecting the training convergence as some non-negligible gradient signals were skipped. 

\begin{figure}[h]
    \centering
    \includegraphics[width=0.65\textwidth]{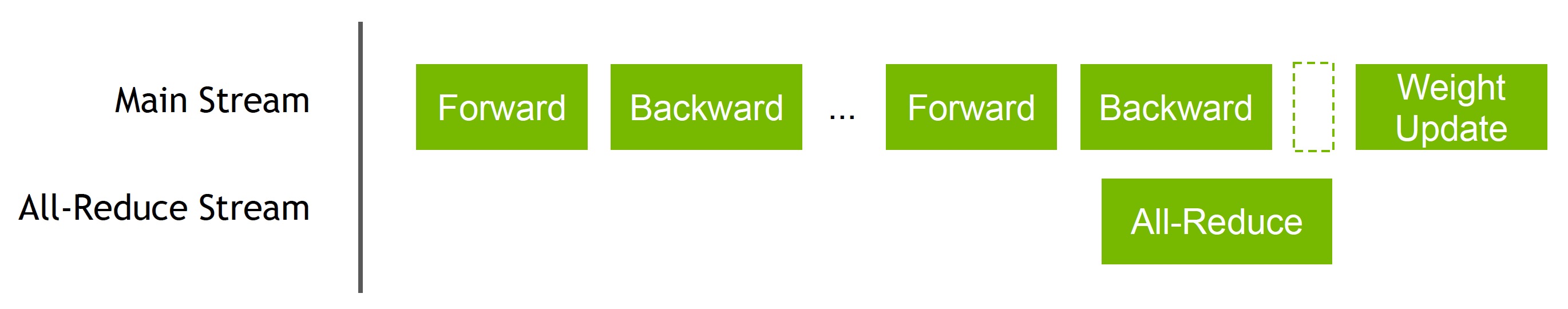}
    \caption{CUDA Stream Timeline for Gradient Accumulated Training}
    \label{fig:gradient-accumulated-nccl-timeline}
\end{figure}

To address the aforementioned communication bottleneck, gradient accumulation is applied. Gradient accumulation is the process of adding up the loss and gradients in each local worker over multiple mini-batch iterations and updating the weights of the model globally once in every several iterations. As Figure~\ref{fig:gradient-accumulated-nccl-timeline} illustrates, gradient accumulation essentially reduces the ratio between communication workload and computation workload. Since our hardware setup is network bandwidth limited, a properly tuned gradient accumulation step can effectively balance the computation and communication time, and therefore increases the overall compute utilization rate. Note that gradient accumulation also effectively increases the batch size of the training, so other hyper-parameters need to be adjusted accordingly. 

\section{Evaluation}
We describe our performance improvements for the different layers of optimizations below.
\subsection{Single GPU Optimization}
For the single kernel optimization, Table~\ref{tab:table1} and \ref{tab:table2} summarize the throughput gain of using FP16 and kernel fusion. Using FP16 improves the throughput by 1.7$\times$  on NVIDIA P100 and 2.5$\times$ on NVIDIA 2080 Ti. Furthermore, FP16 is more effective on GPUs equipped with TensorCores as the cores are enabled only by FP16 operations. Kernel fusion further enhances the single-GPU throughput by around 1.2$\times$ for all three devices. Since these optimization techniques can be applied separately, the combination of both produces a final speed up of at least 2.05$\times$ on NVIDIA P100, 2.78$\times$ on NVIDIA T4, and 3.05$\times$ on NVIDIA 2080 Ti.

\begin{table}[hbt!]
 \caption{Throughput Comparison (Tokens/s)}
  \centering
  \begin{tabular}{lllll}
    \toprule
    Device              & Non-Optimized     & FP16              & FP16 \& Fused Kernel  & Seq Length\\
    \midrule
    P100~\cite{nvp100}               & 1576.3              & 2680.7          & 3228.8    & 128\\
    T4 (TensorCore)~\cite{nvt4}      & 1953.5             & 4430.9          & 5429.1   & 128\\
    2080Ti (TensorCore)~\cite{nvrtx2080}   &   3527.2            &  8823.8         & 10765.8   & 128\\
    \bottomrule
  \end{tabular}
  \label{tab:table1}
\end{table}

\begin{table}[hbt!]
 \caption{Throughput Speedups (using non-optimized baseline)}
  \centering
  \begin{tabular}{llll}
    \toprule
    Device              & Non-Optimized     & FP16              & FP16 \& Fused Kernel \\
    \midrule
    P100~\cite{nvp100}                & 1              & 1.7            & 2.05\\
    T4 (TensorCore)~\cite{nvt4}    & 1              & 2.27           & 2.78\\
    2080Ti (TensorCore)~\cite{nvrtx2080}  & 1              & 2.5            & 3.05\\
    \bottomrule
  \end{tabular}

  \label{tab:table2}
\end{table}

\subsection{Multi-Node Optimization}

\begin{figure}[h]
    \centering
    \includegraphics[width=0.7\textwidth]{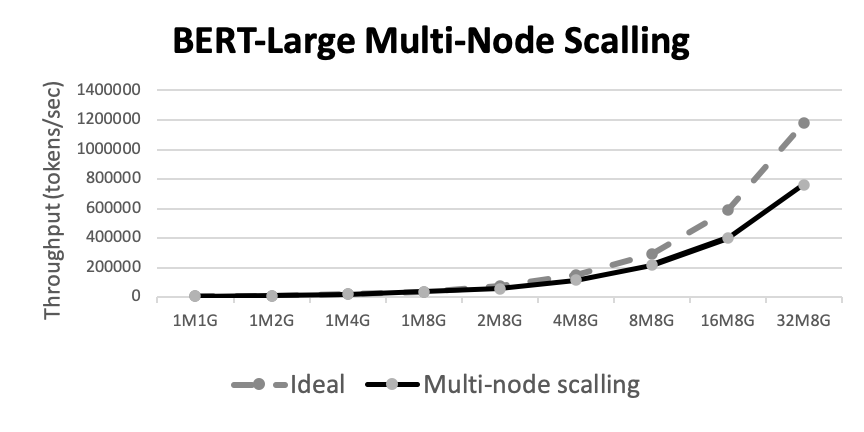}
    \caption{Multi-node Throughput Scaling}
    \label{fig:multinode}
\end{figure}

We trained BERT-large with 32 machine-nodes, each equipped with 8 NVIDIA T4 GPUs ~\cite{nvt4}. This amounts to 256 GPUs in total. We applied gradient accumulation for 4 steps to reduce network traffic. Individual GPU workers sum up the gradients from 4 different mini-batches before exchanging and updating the model parameter among all the workers. Combining the reduction of network traffic from performing gradient accumulation with our optimization work on single GPU, we are able to achieve a weak scaling factor of 165 times with 10 Gbps network bandwidth. As we show in Figure~\ref{fig:multinode}, the scaling efficiency decreases as we continue to increase the number of machines as communication and synchronization overhead dominates the training time.

\begin{figure}[h]
    \centering
    \includegraphics[width=0.495\textwidth]{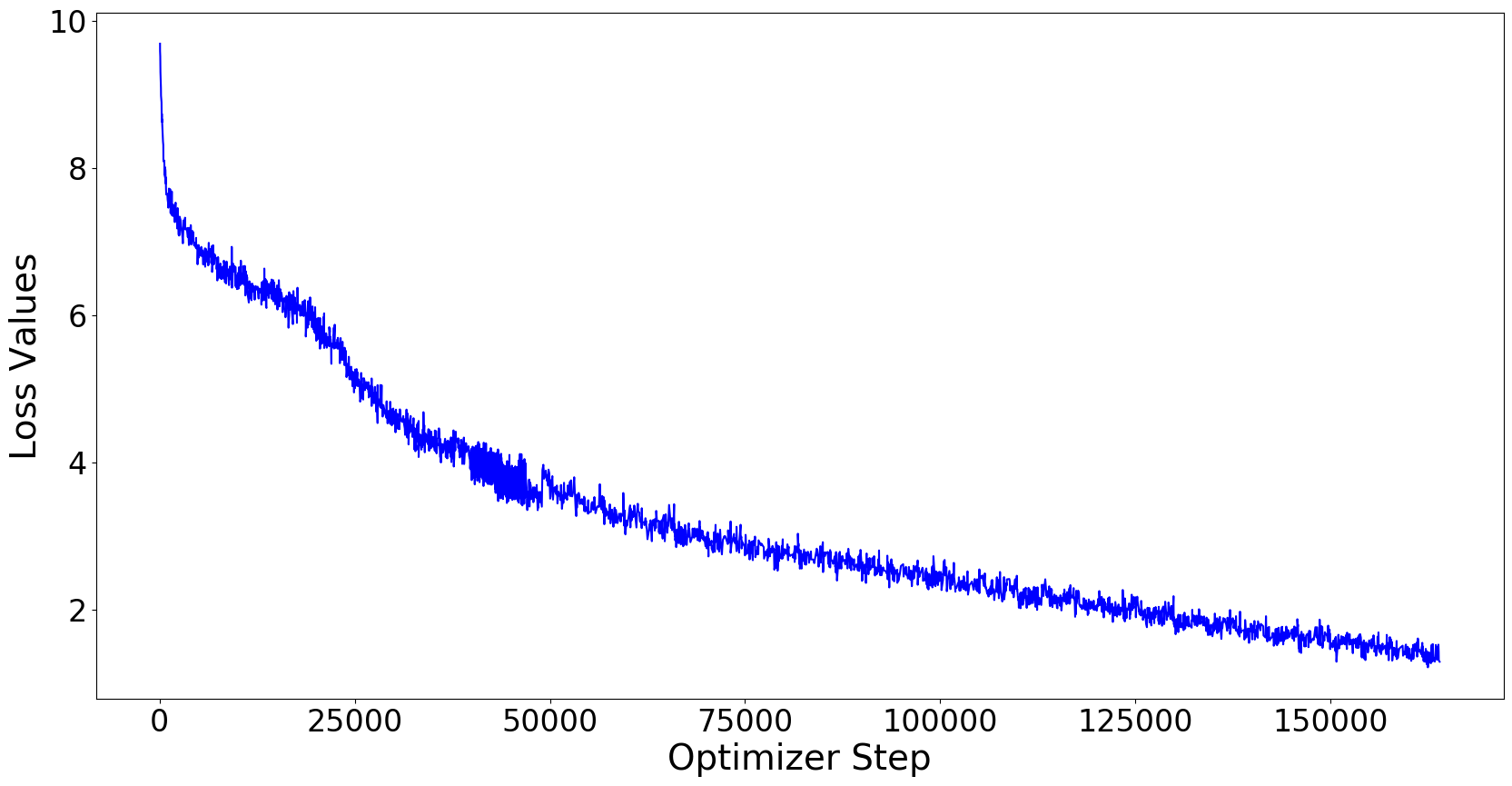}
    \includegraphics[width=0.495\textwidth]{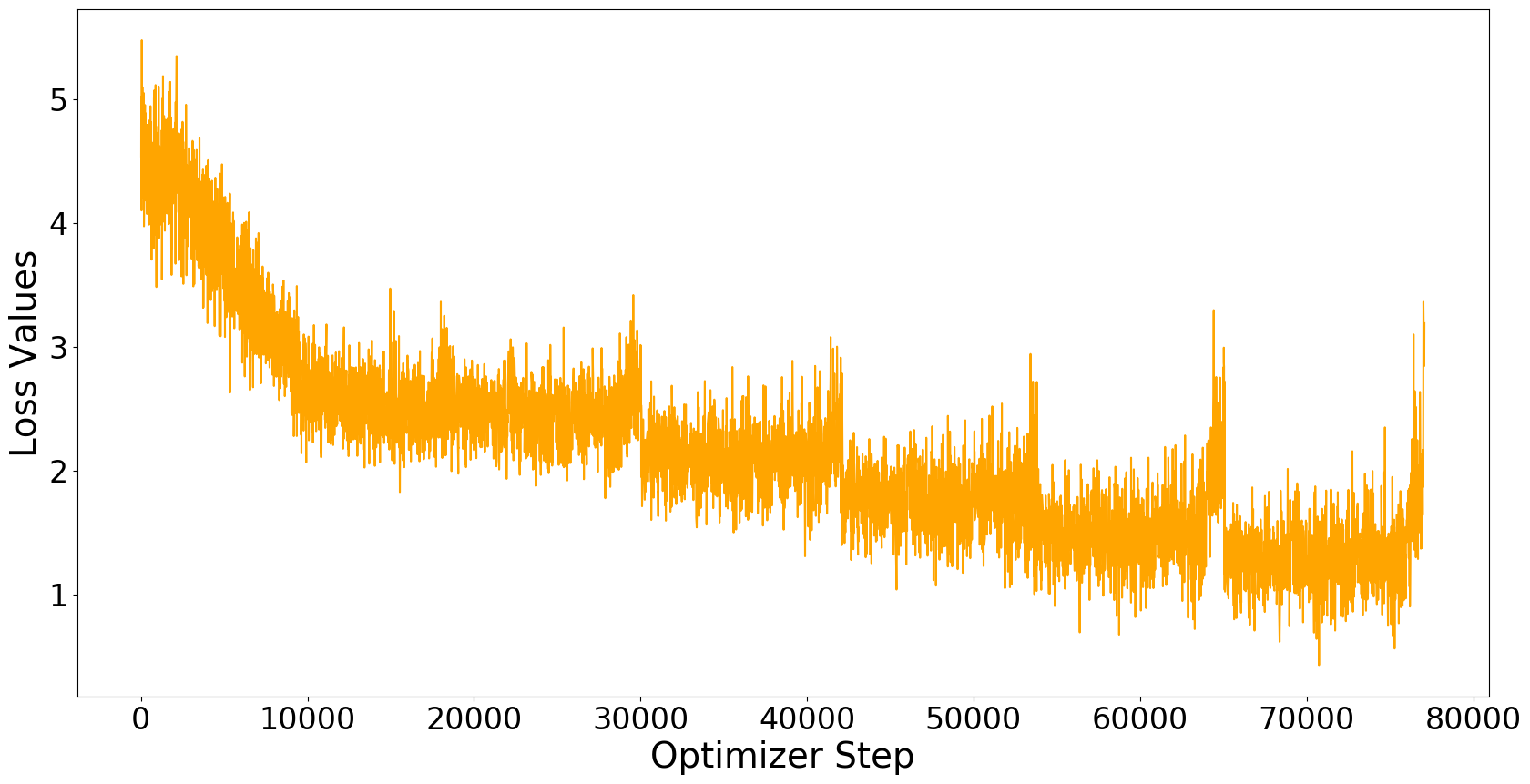}
    \caption{BERT-large Phase 1 (left) \& Phase 2 (right) Pre-training Loss Plot}
    \label{fig:pretrainloss}
\end{figure}

Figure \ref{fig:pretrainloss} shows the loss curves of two phase training and table \ref{tab:table4} listed the differences in training configurations for our two phase pre-training. We had some convergence issues in phase 2, as figure \ref{fig:pretrainloss} illustrates, the training loss plateaus after one epoch of training, and starting from the second epoch, loss spikes up at the very end of each epoch and decreases later. In phase 1 the loss value at the end of last epoch is about 1.2. In phase 2, the average loss value in the final epoch is around 1.3. 

\begin{table}[]
    \caption{Two Phase Pre-training Comparison (per GPU)}
      \centering
      \begin{tabular}{llllllll}
        \toprule
                     & Sentences (S) & Length/S  & Predictions/S  & Batch Size    & Learning Rate & Epochs & Epoch Time\\
        \midrule
        Phase 1         & 32    & 128       & 20   & 4096    & 1e-4     & 36  & 6 hours\\
        Phase 2         & 4     & 512       & 80   & 2048    & 1e-4     & 6  & 16 hours\\
        \bottomrule
      \end{tabular}
      \label{tab:table4}
\end{table}

\begin{figure}[h]
    \centering
    \includegraphics[width=0.495\textwidth]{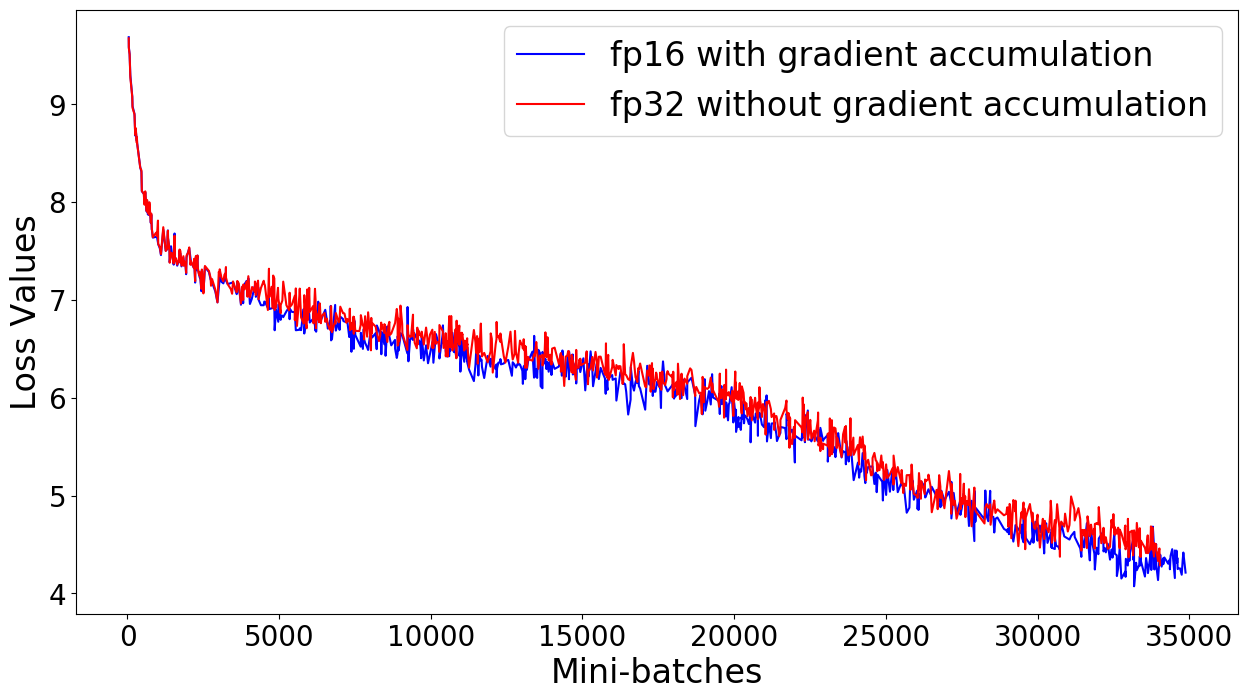}
    \includegraphics[width=0.495\textwidth]{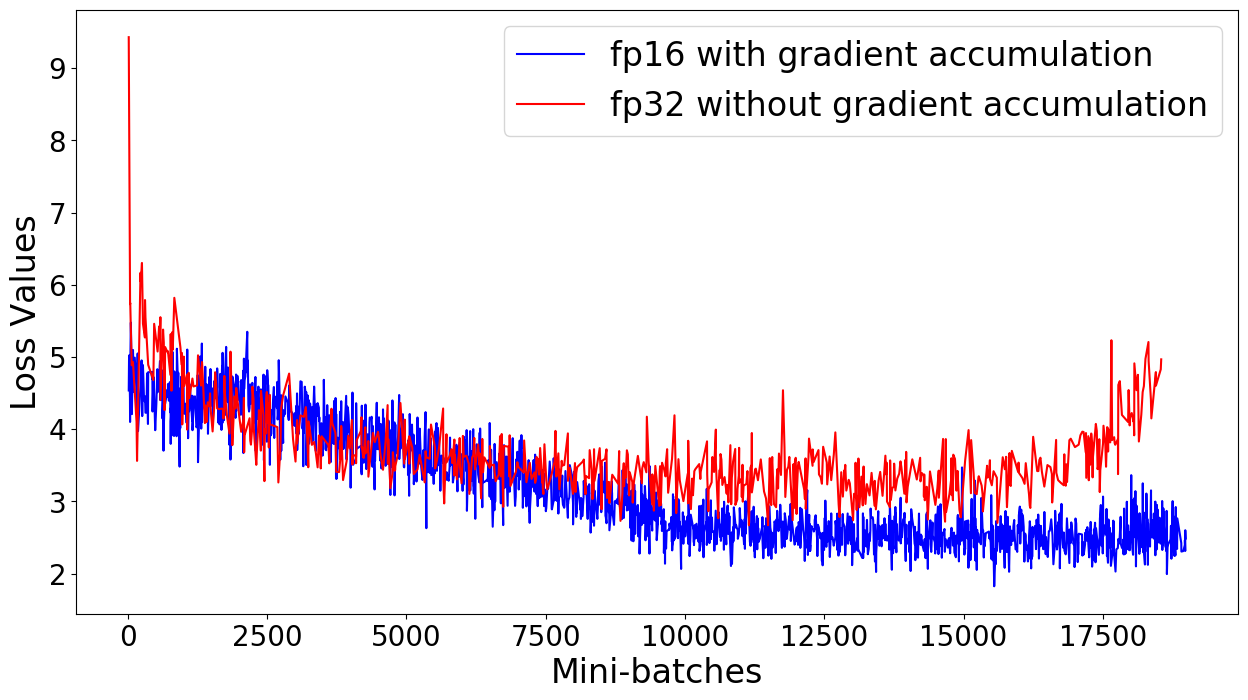}
    \caption{BERT-large Phase 1 (left) \& Phase 2 (right) Optimized vs. Non-optimized Training Loss Comparison}
    \label{fig:losscomparison}
\end{figure}

\subsection{Pre-training and Fine-tuning Results}
We evaluated our pre-training models through fine-tuning our pre-trained model on SQuAD v1.1 dataset using the same fine-tuning configurations as \cite{bert} did. Our model achieved 81\% to 83\% F1 scores depending on the loaded pre-trained checkpoints. Compared with Google's 90.9\% ~\cite{bert} and NVIDIA's ~\cite{bertimplementation} 90\% to 91\%,  there is a discrepancy of 9\%--10\%. 

However, we believe that such discrepancy is not caused by the system performance optimization approach that we had been taken to shorten the pre-training time of BERT-large. To illustrate this, we experimented two sample pre-training run for each phase, one run with the performance optimizations while the other without any of the performance optimization. As Figure~\ref{fig:losscomparison} shows, in phase 1 the two loss curve is highly similar and in phase 2 the optimized loss curve looks even more stable than the non-optimized one. This proves that the convergence issue that we saw in phase 2 might be caused by incorrect hyper-parameter settings. However, to fine-tuning of these hyper-parameters requires excessive amount of computing resources as well as time. Note that we trained extra two epochs in phase 2 to get the desired loss value because of the convergence problem, this makes our total training time 13 days. With an ideal parameter setting, phase 2 training should be completed by 4 epochs and the total training time can be further shortened to be within 12 days.

\section{Conclusion \& Future Work}
We have successfully completed the pre-training of BERT-large in 12 days with a relatively low hardware capital budget than most of the published results by major software/hardware companies~\cite{bert,roberta}. For example, our capital cost for the experiment is estimated to be \$624K for a total of 32 nodes, where as the DGX workstations used for NVIDIA's pre-training of BERT cost around \$4.8M to \$12.8M. The choice of owning the hardware is also more cost efficient than renting is through major cloud service providers. For example, the cost of 256 T4 GPUs on Google Cloud for 12 days is estimated to be \$25739 (Appendix \ref{appendix:graph}), which is 24 times less the price of owning the hardware. However, the average replacement cycle for the hardware is about 3 years, which is enough time for 90 times of such a 12-day experiment.  

\section*{Acknowledgements}
We want to thank Vector Institute NLP Project industry and technical staff participants who gave feedback. We want to offer special thanks to Dr. Garth Gibson and Dr. Elham Dolatabadi from Vector Institute for their guidance and support throughout this work, Punendu Mukherjee and Thor Johnsen from NVIDIA, Fillippo Pompilli from Thomson Reuters for their help during preliminary phase of experiments.

This work was supported in part by the NSERC Discovery grant, the Canada Foundation for Innovation JELF grant, the Connaught Fund, the Huawei grants, the Province of Ontario, the Government of Canada through CIFAR AI Chair award, and sponsors of the Vector Institute (\url{www.vectorinstitute.ai/\#partners}).

\bibliographystyle{unsrt}
\bibliography{main} 

\newpage
\section*{Appendices}
\subsection*{Training Cost Estimation}

\label{appendix:graph}
\begin{table}[h]
    \caption{Google Cloud Price Estimation}
      \centering
      \begin{tabular}{lclll}
        \toprule
        Devices     & Number of Devices & Price/hour (USD)  & Traing Time  & Total Cost (USD)\\
        \midrule
        NVIDIA T4     & 256    & \$0.35 ~\cite{google_cloud_gpu}       & 12 Days   & \$25804.8  \\
        \bottomrule
      \end{tabular}
      \label{tab:table4}
\end{table}

\begin{table}[h]
    \caption{NVIDIA DGX Cluster Price Estimation}
      \centering
      \begin{tabular}{lclll}
        \toprule
        Devices     & Number of Devices & Price (USD)  & Total Cost (USD)\\
        \midrule
        NVIDIA DGX1   &32 & \$149,000 ~\cite{dgx_price}         & \$4768000  \\
        NVIDIA DGX2   &32 & \$399,000 ~\cite{dgx_price}   & \$12768000  \\
        \bottomrule
      \end{tabular}
      \label{tab:table4}
\end{table}

\end{document}